\documentclass[conference]{IEEEtran}
\IEEEoverridecommandlockouts

\usepackage{cite}

\usepackage{amsmath,amssymb,amsfonts}
\usepackage{graphicx}
\usepackage{textcomp}
\usepackage{xcolor}
\usepackage{array}
\usepackage{tabularx}
\usepackage{multicol}
\usepackage{bm}
\usepackage{booktabs}
\usepackage{float}
\usepackage{algorithm}
\usepackage{algpseudocode}
\usepackage{subfigure}

\usepackage[colorlinks=true]{hyperref}
\definecolor{citegreen}{rgb}{0.2,0.8,0.2}
\definecolor{refered}{rgb}{1.0,0.0,0.0}
\hypersetup{
    citecolor=citegreen,    
    linkcolor=refered,        
    urlcolor=magenta,          
    filecolor=black        
}

\algrenewcommand\algorithmicrequire{\textbf{Input:}}
\algrenewcommand\algorithmicensure{\textbf{Output:}}
\def\BibTeX{{\rm B\kern-.05em{\sc i\kern-.025em b}\kern-.08em
    T\kern-.1667em\lower.7ex\hbox{E}\kern-.125emX}}
\begin{document}
\title{JojoSCL: Shrinkage Contrastive Learning for single-cell RNA sequence Clustering*\\
}
\author{\IEEEauthorblockN{1\textsuperscript{st} Ziwen Wang}
\IEEEauthorblockA{\textit{Department of Computer Science} \\
\textit{New York University, Courant Institute of Mathematical Sciences}\\
New York City, U.S. \\
zw1663@nyu.edu}
}

\maketitle

\begin{abstract}
Single-cell RNA sequencing (scRNA-seq) has revolutionized our understanding of cellular processes by enabling gene expression analysis at the individual cell level. Clustering allows for the identification of cell types and the further discovery of intrinsic patterns in single-cell data. However, the high dimensionality and sparsity of scRNA-seq data continue to challenge existing clustering models. In this paper, we introduce JojoSCL, a novel self-supervised contrastive learning framework for scRNA-seq clustering. 
By incorporating a shrinkage estimator based on hierarchical Bayesian estimation, which adjusts gene expression estimates towards more reliable cluster centroids to reduce intra-cluster dispersion, and optimized using Stein’s Unbiased Risk Estimate (SURE), JojoSCL refines both instance-level and cluster-level contrastive learning. Experiments on ten scRNA-seq datasets substantiate that JojoSCL consistently outperforms prevalent clustering methods, with further validation of its practicality through robustness analysis and ablation studies. JojoSCL's code is available at: \url{https://github.com/ziwenwang28/JojoSCL}.
\end{abstract}

\begin{IEEEkeywords}
ScRNA-seq Clustering, Contrastive  Learning, Bayesian hierarchical modeling, Shrinkage estimator
\end{IEEEkeywords}

\section{Introduction}
The advancement of single-cell RNA sequence (scRNA-seq) technology has driven breakthroughs in various fields including developmental biology, cancer research, and precision medicine \cite{b1}, and the accurate identification of cell types enables the further analysis in their functions and dynamics and deepens our understanding of the cellular biology and disease mechanisms \cite{b2}. Clustering is a powerful method for cell type identification through examining structural similarities and differences between cells \cite{b3}. Early methods, such as CIDR \cite{b4} and SIMLR \cite{b5}, addressed the challenges of high dimensionality and sparsity in scRNA-seq data through statistical techniques in dimensionality reduction.
To strengthen clustering robustness, ensemble clustering approaches were introduced. These methods aggregate clustering results into a consensus matrix, with SAFE \cite{b6} synthesizing outputs from various procedures, including t-SNE, CIDR, and Seurat \cite{b7}, to improve clustering performance by integrating a broad range of data characteristics. Nonetheless, these early approaches cannot fully capture the complex structures in scRNA-seq data.

\begin{figure*}[t]
    \centering
    \includegraphics[width=\linewidth, height=0.30\textheight]{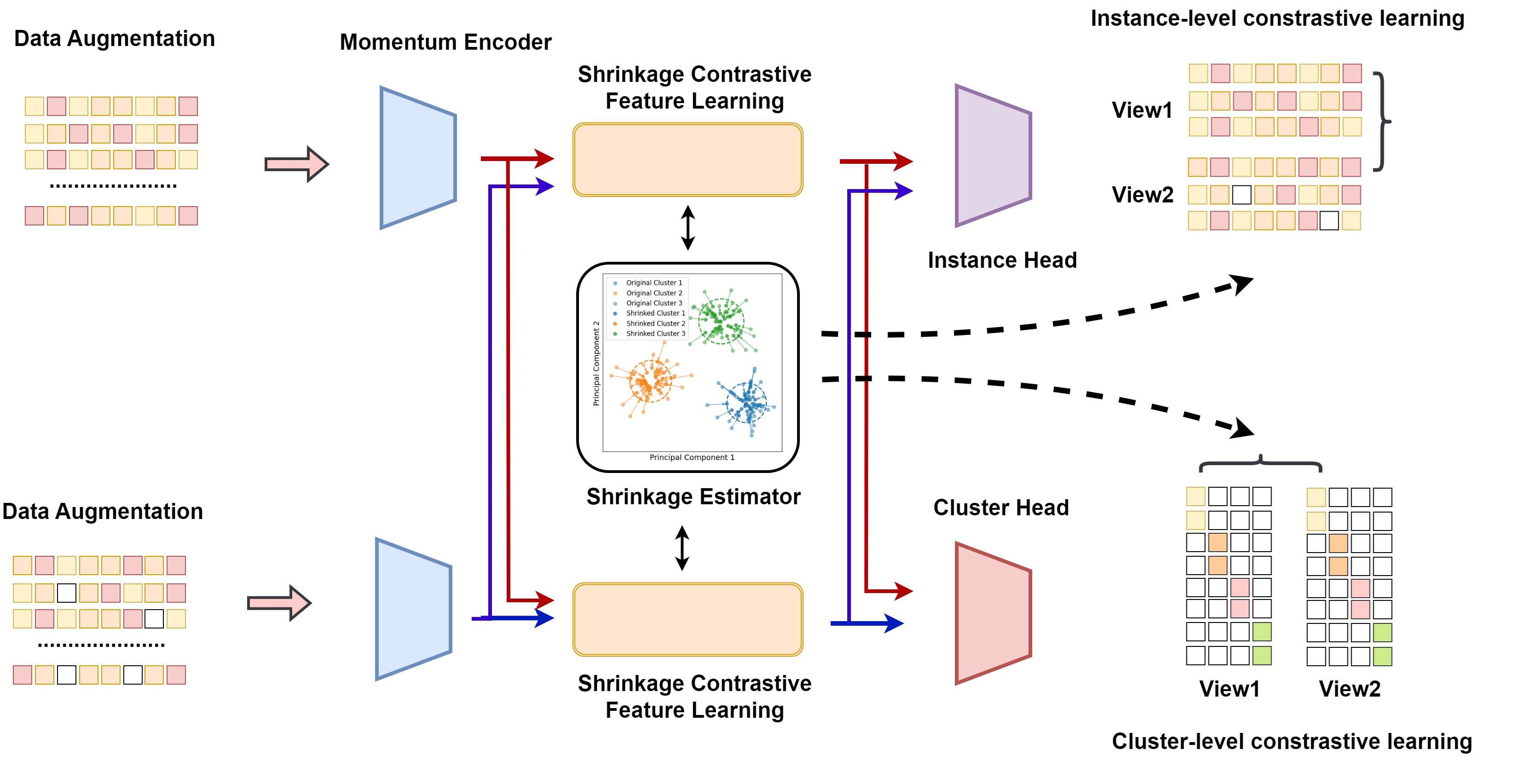}
    \caption{The overview of the proposed model. The model consists of data augmentation and shrinkage contrastive learning. Our approach utilizes a shrinkage estimator to enhance both instance-level and cluster-level contrastive loss functions (depicted by the dashed line). Collectively, the instance-level loss, cluster-level loss, and shrinkage feature learning components form a comprehensive loss function that effectively guides the training process.}
    \label{fig: JojoSCL}
\end{figure*}

Recent advancements in deep learning \cite{b8,b9} have led to the development of deep clustering methods for identifying scRNA-seq cell types. Deep neural networks (DNNs) facilitate nonlinear dimensionality reduction, thereby enhancing clustering performance \cite{b10}. Notably, DeepImpute \cite{b11} adopts DNNs to predict missing values by exploiting gene correlations, while DCA \cite{b12} utilizes a zero-inflated negative binomial (ZINB) model to address dropout events and improve data reconstruction. Further progress is exemplified by scDeepCluster \cite{b13}, which integrates DCA with the Deep Embedded Clustering (DEC) algorithm to achieve dimensionality reduction and clustering within a unified framework, which then optimizes clustering outcomes through simultaneous learning from a low-dimensional data representation.

Contrastive clustering \cite{b14} stems from contrastive learning 
\cite{b15,b16} by maximizing the similarity between similar embeddings and minimizing it between dissimilar ones, consequently improving representation quality and cluster separation. It has been adapted for scRNA-seq data due to its efficacy in capturing meaningful features in high-dimensional data. Contrastive-sc \cite{b17} improves distinction between similar and dissimilar cells by masking random features to create augmented pairs as positive samples. scNAME \cite{b18} advances contrastive learning through neighborhood contrastive loss and mask estimation to better capture feature correlations and pairwise similarities using local neighborhood information. CLEAR \cite{b19} leverages multiple data augmentations and the infoNCE \cite{b20} loss with momentum updates to address noise and refine feature representations \cite{b21}. ScCCL \cite{b22} integrates gene expression masking, Gaussian noise, and a momentum encoder \cite{b23} to obtain high-order embeddings and uses a loss function that combines instance- and cluster-level contrastive learning.

Despite contrastive learning's success in scRNA-seq clustering and feature discovery, traditional data augmentation and feature extraction methods still often fail to address the data's inherent sparsity and noise. Applying these methods without addressing these issues can exacerbate the learning of erroneous data distributions.  Instead, guiding the learning process toward well-defined centroids in high-dimensional space \cite{b24} can reduce dispersion and increase information precision. We propose JojoSCL, a novel contrastive clustering framework with a jointly optimized shrinkage scheme. Inspired by the James-Stein (JS) estimator, a biased shrinkage estimator used in higher-dimensional statistics, we implement a hierarchical Bayesian distribution model \cite{b25}, constrained by Stein's Unbiased Risk Estimate (SURE) 
\cite{b26,b27} as a loss function, that simultaneously accounts for the variability in individual data points and the uncertainty in the true cluster centroids by integrating prior information at multiple levels to effectively capture the structure of the scRNA-seq data and refine the estimates of both the cluster centroids and the distribution of data points within each cluster. Our refinement also contributes to the ongoing analysis regarding the effectiveness of contrastive learning
\cite{b28,b29} with more negative samples \cite{b30} and the corresponding methods for generating them \cite{b31}. By focusing on enhancing intra-cluster concentration leveraging bias-variance tradeoff, we improve the quality of both positive and negative samples used in the learning process to strengthen the contrastive learning framework.
The primary contributions of JojoSCL are:

\noindent 1. We introduce an innovative self-supervised contrastive learning framework that incorporates a novel shrinkage estimator. It effectively addresses the challenges posed by high dimensionality and sparsity in scRNA-seq data for scRNA-seq clustering.

\noindent 2. We demonstrate that this shrinkage strategy elevates both instance-level and cluster-level learning in contrastive learning.

\noindent 3. Empirical results across ten scRNA-seq datasets demonstrate that JojoSCL significantly surpasses the benchmark models. Furthermore, an in-depth analysis of the model confirms the robustness and efficacy of our model.

\section{Shrinkage Estimator}

We assume scRNA-seq data points are normally distributed around cluster centroids. Despite variability in gene expressions within the same population, we assume that each gene's expression shares a common variance across those cells, with a consistent covariance matrix across clusters. For inter-cluster gene expressions, we use a multivariate normal distribution for cluster centroids, with parameters estimated from an informed prior distribution. This section details the mathematical formulation of our framework, addressing the limitations of traditional methods with high-dimensional, sparse data. Building on the intuition of the JS estimator, we introduce a hierarchical Bayesian model to better capture scRNA-seq data structure, refine centroid estimates, and optimize clustering performance.

\subsection{Preliminary: Theoretical Background}

Let $\bm{X}$ be a $P$-dimensional vector representing an scRNA-seq data point, where each component of $\bm{X}$ is an independent and identically distributed (i.i.d.) normal random variable with unknown mean $\theta$ and known common variance $\sigma^2$. The Maximum Likelihood Estimator (MLE) for \(\theta\) is given as
\(
\theta_{\text{MLE}} = \bm{X}.
\)
The James-Stein (JS) estimator shrinks the estimates towards the origin and lower the mean squared error (MSE) in high-dimensional contexts \cite{b32}. For \(\bm{X} \sim \text{Normal}(\theta, \sigma^2)\) with the dimension-specific parameters \(\left\{ \theta_p \mid p=1\ldots P \right\} \) and \(\left\{ \sigma_p^2 = \sigma^2 \mid p=1\ldots P \right\} \), the JS estimator is:
\[
\scalebox{1}{$
\theta_{\text{JS}} = \left(1 - \frac{(P - 2)\sigma^2}{\|\bm{X}\|_2^2} \right) \bm{X}
$}
\tag{1}
\label{1}
\]
The MSE of the JS estimator is:
\[
\scalebox{1}{$
\text{MSE}(\theta_{\text{JS}}) = \text{MSE}(\theta_{\text{MLE}}) - {(P - 2)^2 \sigma^4}{E\left[\frac{1}{\|\bm{X}\|_2^2}\right]},
$}
\label{2}
\tag{2}
\]
where the second term is defined and positive for \( P \geq 3 \), suggesting that \(\theta_{\text{JS}}\) is guaranteed to have a lower MSE than \(\theta_{\text{MLE}}\) in higher dimensions. Despite potential higher MSE in specific dimensions of \(\theta\), particularly when \(\theta_p\) is distant from the origin, the much more significant reduction in overall MSE benefits high-dimensional settings like scRNA-seq data by minimizing noise and enhancing estimate precision.

\subsection{Parameter Estimation and Derivation of Key Equations}

The \(\theta_{\text{MLE}}\) estimator maximizes the likelihood function based solely on observed data. In contrast, \(\theta_{\text{JS}}\) leverages an empirical Bayesian approach, estimating the prior distribution from the data [33, 34]. This is known as Maximum A Posteriori (MAP) estimation:
\[
\scalebox{1}{$
\hat{\theta}_{\text{MAP}} = \arg \max_{\theta} \pi_{\Theta}(\theta \mid \bm{x}) = \arg \max_{\theta} L(\bm{x} \mid \theta) \pi_{\Theta}(\theta),
$}
\tag{3}
\]
where \(\pi_{\Theta}(\theta \mid \bm{x})\) is the posterior distribution, \(L(\bm{x} \mid \theta)\) the likelihood function, and \(\pi_{\Theta}(\theta)\) the prior. While \(\theta_{\text{JS}}\) assumes \(\theta\) is near the origin \(\mathbf{0}_P\) (see (\ref{1})), this assumption can be ineffective for scRNA-seq data, where \(\mathbf{0}_P\) may not accurately represent the true cluster mean. To better model scRNA-seq data, we assume each component \(\bm{X}_p\) follows:
\[
\scalebox{1}{$
\bm{X}_p \sim \text{Normal}(\theta_p, \sigma^2),
$}
\tag{4}
\]
and the prior distribution of \(\theta_p\) is:
\[
\scalebox{1}{$
\theta_p \sim \text{Normal}(\mu_p, \tau^2).
$}
\tag{5}
\]
Thus, we can derive the Maximum A Posteriori (MAP) estimate \(\hat{\theta}_{\text{MAP}}\) of \(\theta\) by integrating the information provided by the hierarchical observation \(\bm{X}\) with the assumed prior distributions. This is achieved by maximizing the posterior distribution:
\[
\scalebox{0.95}{$
\begin{aligned}
\hat{\theta}_{\text{MAP}} &= \arg\max_{\theta} \left(f_{\bm{X}|\theta}(\bm{X} \mid \theta) f_\theta(\theta)\right) \\
&= \arg \max_\theta \left(\frac{1}{\sqrt{2\pi \sigma^2}} \exp\left(-\frac{(\bm{X} - \theta)^2}{2\sigma^2}\right)\right. \\
&\quad \left. \cdot \frac{1}{\sqrt{2\pi \tau^2}} \exp\left(-\frac{(\theta - \mu)^2}{2\tau^2}\right)\right).
\end{aligned}
$}
\tag{6}
\]
Solving (6), we obtain the MAP estimator:
\[
\scalebox{1}{$
\hat{\theta}_{\text{MAP}} (\sigma, \bm{X}; \mu, \tau) = \frac{\tau^2}{\tau^2 + \sigma^2} \bm{X} + \frac{\sigma^2}{\tau^2 + \sigma^2} \mu.
$}
\tag{7}
\label{7}
\]

To determine the distribution of \( \bm{X} \) within the hierarchical framework where \( \bm{X} \sim \text{Normal}(\theta, \sigma^2) \) and \( \theta \sim \text{Normal}(\mu, \tau^2) \), we calculate its compound probability distribution by integration:
\[
\scalebox{1}{$
\begin{aligned}
f_{\bm{X}}(\bm{x}) &= \int_{-\infty}^{\infty} f_{\bm{X}|\theta}(\bm{x} \mid \theta) \, f_{\theta}(\theta) \, d\theta \\
&= \frac{1}{2\pi (\sigma^2 + \tau^2)} \exp\left( -\frac{(\bm{x} - \mu)^2}{2(\sigma^2 + \tau^2)} \right).
\end{aligned}
$}
\tag{8}
\label{8}
\]
\[
\scalebox{1}{$
\bm{X} \sim \text{Normal}(\mu, \sigma^2 + \tau^2).
$}
\tag{9}
\]
This result supports the assertion that the MAP estimator \(\hat{\theta}_{MAP}(\sigma, \bm{X}; \mu, \tau)\), as derived in (\ref{8}), achieves a guaranteed lower mean squared error (MSE) compared to the MLE estimator. Specifically:
\[
\scalebox{0.97}{$
\text{MSE}(\hat{\theta}_{\text{MLE}}) = \mathbb{E}\left[\left\| \bm{X} - \mu \right\|_2^2\right]
$}
\]
\[
\scalebox{0.97}{$
\begin{aligned}
\text{MSE}\left( \hat{\theta}_{MAP} \right) &= \mathbb{E}\left[\left\| \frac{\tau^2}{\tau^2 + \sigma^2}\bm{X} + \frac{\sigma^2}{\tau^2 + \sigma^2}\mu - \mu \right\|_2^2\right] \\
&= \mathbb{E}\left[\left\| \frac{\tau^2}{\tau^2 + \sigma^2}(\bm{X} - \mu) \right\|_2^2\right]
\end{aligned}
$}
\tag{10}
\]
where \({\tau^2}/{(\tau^2 + \sigma^2)} \leq 1\). Given that \(\hat{\theta}_{\text{MAP}}\) is a biased and nonlinear estimator of \(\theta\) with respect to the parameters \(\mu\) and \(\tau\), which must be estimated, Stein's Unbiased Risk Estimate (SURE) can be used to provide an unbiased estimate of the MSE of \(\hat{\theta}_{\text{MAP}}\). The SURE of \(\hat{\theta}_{MAP} (\sigma, \bm{X}; \mu, \tau)\) is given by:
\[
\scalebox{0.97}{$
\begin{aligned}
\text{SURE}(\hat{\theta}_{MAP} (\sigma, \bm{X}; \mu, \tau)) &= -P\sigma^2 + \|\hat{\theta}_{MAP} - \bm{X}\|_2^2 \\
&\quad + 2\sigma^2 \sum_{p=1}^{P} \frac{\partial \hat{\theta}_{MAP} (\sigma, \bm{X}; \mu, \tau)^T}{\partial \bm{X}_p}.
\end{aligned}
$}
\tag{11}
\label{11}
\]
By substituting the MAP estimator \(\hat{\theta}_{MAP} (\sigma, \bm{X}; \mu, \tau)\) defined in (\ref{7}) into (\ref{11}), we can simplify the expression:
\[
\scalebox{0.85
}{$
\begin{aligned}
\text{SURE}(\hat{\theta}_{\text{MAP}} (\sigma, \bm{X}; \mu, \tau)) 
& = -P\sigma^2 + \left\|\frac{\tau^2}{\tau^2 + \sigma^2} \bm{X} + \frac{\sigma^2}{\tau^2 + \sigma^2} \mu - \bm{X}\right\|_2^2 \\
& \quad + 2\sigma^2 \sum_{p=1}^P \frac{\partial \left(\frac{\tau^2}{\tau^2 + \sigma^2} \bm{X} + \frac{\sigma^2}{\tau^2 + \sigma^2} \mu\right)^T}{\partial \bm{X}_p}
\end{aligned}
$}
\]
\[
\scalebox{0.88}{$
\begin{aligned}
& = -P\sigma^2 +  \left\| \frac{-\sigma^2}{\tau^2+\sigma^2}\bm{X} + \frac{\sigma^2}{\tau^2+\sigma^2}\mu \right\|_2^2 + 2\sigma^2 \cdot P \cdot \frac{\tau^2}{\tau^2+\sigma^2}\\
& = \frac{\sigma^2}{\tau^2+\sigma^2} \|\mu - \bm{X}\|_2^2 + P\sigma^2 \left(\frac{\tau^2 - \sigma^2}{\tau^2+\sigma^2}\right)\\
&= \frac{\sigma^2}{\tau^2+\sigma^2} \left( \|\mu-\bm{X}\|_2^2 + P\left(\tau^2 - \sigma^2\right) \right),
\end{aligned}
$}
\tag{12}
\label{12}
\]
which provides an unbiased estimate of the risk associated with \(\hat{\theta}_{MAP} (\sigma, \bm{X}; \mu, \tau)\).

With \(\mu\) and \(\tau\) derived from the prior distribution, we estimate these parameters from the data and subsequently compute the corresponding \(\text{SURE}(\hat{\theta}_{MAP} (\sigma, \bm{X}; \mu, \tau))\). By selecting the pairs \((\mu, \tau)\) that minimize \(\text{SURE}(\hat{\theta}_{MAP} (\sigma, \bm{X}; \mu, \tau))\), we obtain the optimized parameters for describing the prior distributions, which can then be substituted back into (\ref{7}) to compute the shrinkage estimator \(\hat{\theta}_{MAP}\) for \(\bm{X}\). Thus, this procedure ensures that \(\hat{\theta}_{MAP}\) is optimized for the data point \(\bm{X}\).

\subsection{Shrinkage Estimator on scRNA-seq Clustering Tasks}

We extend the MSE minimization to multiple centroids \(\left\{ C_k \mid k = 1,\ldots,K \right\}\) using hierarchical Bayesian inference to better align observations with centroids. The effectiveness of this method is evaluated by the aggregate \(\text{SURE}(\hat{\theta}_{\text{MAP}} (\sigma, \bm{X}; \mu, \tau))\) over all data points, with lower aggregate SURE values indicating better overall MSE reduction. This shrinkage framework boosts contrastive learning by focusing on discriminative features and reducing the influence of variable genes, thus improving the model's ability to differentiate between similar and dissimilar cells. Details on implementation are provided in Section 3.

\section{Contrastive Clustering with Shrinkage Estimator}

This section presents the implementation of JojoSCL, which uses a momentum-based encoder to stabilize the learning process and utilizes SURE optimization to minimize intra-cluster dispersion. The shrinkage estimator refines both instance-level and cluster-level contrastive loss functions, creating a unified loss function that directs the training process. This integrated approach aims to optimize the identification and separation of cell types in scRNA-seq data. An overview of the model is illustrated in Fig. \ref{fig: JojoSCL}.

\subsection{Contrastive Representation Learning}

For contrastive clustering, we apply a data augmentation strategy inspired by ScDeepCluster to scRNA-seq data, which masks some gene expression values and adds Gaussian noise, creating two augmented views, \(\bm{X}_i^a\) and \(\bm{X}_i^b\), from each sample \(\bm{X}_i\). Thus, the sample space expands from \(N\) to \(2N\). Contrastive learning is performed on these views, with \(\{(\bm{X}_i^a, \bm{X}_i^b) \mid i = 1, 2, \ldots, N\}\) as positive pairs and \(\{(\bm{X}_i^a, \bm{X}_j^k) \mid i \neq j \text{ or } k \neq b\}\) as negative pairs, enabling robust feature learning through pairwise comparison.

To address instability due to the high dimensionality and variability in scRNA-seq data, we use a momentum-based encoder framework \cite{b22}. This framework employs two identical encoders, \( f_q \) and \( f_k \), with parameters \( \theta_q \) and \( \theta_k \), respectively. During training, \( \theta_q \) is updated via backpropagation, while \( \theta_k \) is updated with momentum:
\[
\theta_k = m \theta_k + (1 - m) \theta_q,
\tag{13}
\label{13}
\]
where \( m \) is the momentum coefficient. This approach smooths updates for \( \theta_k \), and feature representations \( \bm{h}_{i}^a = f_q(\bm{X}_i^a) \) and \( \bm{h}_{i}^b = f_k(\bm{X}_i^b) \) are obtained from the augmented views.

\subsection{Shrinkage Estimator with SURE Optimization in \texorpdfstring{\(K\)}{K} clusters}

Minimizing the aggregate $\text{SURE}(\hat{\theta}_{MAP} (\sigma, \bm{X}; \mu, \tau))$ effectively aligns data points with their respective centroids, reduces intra-cluster dispersion, and improves the learning process by addressing information loss and noise. However, several issues need to be addressed:

\noindent 1. As outlined in \((\ref{8})\), optimizing data with SURE requires defining the prior distribution for multiple clusters. Specifically, the prior for $\bm{X} \sim \text{Normal}(\theta, \sigma^2)$ is modeled as $\theta \sim \text{Normal}(\mu, \tau^2)$. For $K$ clusters with centroids $\{ C_k \mid k = 1, \ldots, K \}$, a total of $2K$ parameters are needed to describe the distribution of each centroid:
\[
\scalebox{1}{
    $\theta_{k} \sim \text{Normal}(\mu_{k}, \tau_k^2).$
}
\tag{14}
\label{14}
\]

\noindent 2. Accurate estimation of these parameters is essential for identifying $k$ pairs of $\{\mu_k, \tau_k^2\}$ that minimize the aggregate SURE, which will ensure that the clustering model accurately reflects the underlying structure of the data.

To address the first issue, we run the K-means algorithm on the features \( \bm{h}_i^a \) to assign temporal cluster label \(k\) for predicted classification, denoted as \( \left\{ \bm{h}_{i,k}^a \mid k=1,\ldots, K\right\}\). Given \( \bm{h}_{i}^a = f_q(\bm{X}_i^a) \), we can substitute \( \bm{X} \) with \( \bm{h}_{i,k}^a \) in (\ref{12}) to derive the \(\text{SURE}(\hat{\theta}_{MAP} (\sigma, \bm{X}; \mu, \tau)) = \text{SURE}(\hat{\theta}_{MAP} (\sigma_k, \bm{h}_{i,k}^a; \mu_{k}, \tau_k))\) for an individual data point:
\[
\scalebox{1}{$
\text{SURE}(\hat{\theta}_{MAP}) = \frac{\sigma_k^2}{\tau_k^2 + \sigma_k^2} \left(\left\|\mu_{k} - \bm{h}_{i,k}^a\right\|_2^2 + P \left({\tau_k^2 - \sigma_k^2}\right)\right),
$}
\tag{15}
\]
where \(\sigma_k^2\) represents the intra-cluster variance of cluster \(k\).

To resolve the second issue, we use the mean of the intra-cluster samples to approximate \(\mu_{k}\). Since \(\bm{h}_i^a\) is a vector with each component denoted as \( \left\{ \bm{h}_{ip,k}^a \mid p=1,\ldots,P \right\} \), we estimate \(\mu_k\) component-wise with \(\left\{ \mu_{p,k} \mid p=1,\ldots,P \right\}\):
\[
\scalebox{1}{
\(\hat{\mu}_{p,k} = \frac{1}{N_k} \sum_{i=1}^{N_k} \bm{h}_{ip,k}^a = {\overline{\bm{h}}_{ip,k}^a}\),
}
\tag{16}
\]
where \(N_k\) denotes the number of samples in cluster \(k\), and the Central Limit Theorem (CLT) can be applied to estimate \(\tau_k\):
\[
\scalebox{1}{$
\hat{\tau}_k^2 = \frac{\sigma_k^2}{N_k}.
$}
\tag{17}
\]
As outlined in Section 2.1, assuming that the scRNA-seq data of the same cell type shares a common variance in their gene expressions, we use the intra-cluster component-wise variance \(\left\{ 
\sigma_{p,k}^2 \mid p=1,\ldots,P \right\}\) to estimate the overall intra-cluster variance \(\sigma_k^2\):
\[
\scalebox{1}{$
\hat{\sigma}_{p,k}^2 = \frac{1}{N_k - 1} \sum_{p=1}^{P} \left( \bm{h}_{ip,k}^a - {\overline{\bm{h}}_{ip,k}^a} \right)^2
$}
\tag{18}
\label{18}
\]
\[
\scalebox{1}{$
\hat{\sigma}_k^2 = \frac{1}{P}\sum_{p=1}^P \hat{\sigma}_{p,k}^2.
$}
\tag{19}
\label{19}
\]

As a result, we can define the aggregate SURE estimate as:
\[
\scalebox{1}{$
\sum_{i=1}^N \sum_{k=1}^K  \left[ \frac{\hat{\sigma}_k^2}{\hat{\tau}_k^2 + \hat{\sigma}_k^2} \left( \|\hat{\mu}_{k} - \bm{h}_{i,k}^a\|_2^2 + P \left({\hat{\tau}_k^2 - \hat{\sigma}_k^2}\right) \right) \right]
$}
\tag{20}
\label{20}
\]
The temporal assignment of data points allows us to calculate the intra-cluster variances \(\sigma_{p,k}^2\) and \(\sigma_k^2\) using (\ref{18}) and (\ref{19}). These variances are then used in (\ref{20}) to generate a numerical estimation of the aggregate SURE. This estimation is formulated as the SURE loss function in JojoSCL:
\[
\scalebox{1}{$
\begin{aligned}
\mathcal{L}_{SURE} = \sum_{i=1}^N \sum_{k=1}^K 
\mathbf{1}_{\{i \in k\}} &
\left[ 
\frac{\hat{\sigma}_k^2}{\frac{\hat{\sigma}_k^2} {N_k} +\hat{\sigma}_k^2} \left( \left\|{\overline{\bm{h}}_{i,k}^a} - \bm{h}_{i,k}^a \right\|_2^2 \right. \right. \\
&\left. \left. + P \left( \frac{\hat{\sigma}_k^2}{N_k} - \hat{\sigma}_k^2 \right) \right)
\right],
\end{aligned}
$}
\tag{21}
\label{21}
\]
where the indicator function is \(1\) if \(\bm{h}_i^a\) belongs to cluster \(k\) and \(0\) otherwise.
The loss function \( \mathcal{L}_{\text{SURE}} \) imposes a penalty based on the aggregate dispersion of multi-centroids in clustering tasks. As shown in Fig. \ref{fig: JojoSCL}, this shrinkage estimator integrates into contrastive feature learning, promoting embeddings with reduced variance and aligning them with centroids. During training, the values of \(\mathcal{L}_{\text{SURE}}\) are monitored to guide model adjustments. The model parameters are updated based on the minimum observed \(\mathcal{L}_{\text{SURE}}\) value.

\subsection{Instance-level Loss}

We use a two-layer Multilayer Perceptron (MLP), \( g_I(\cdot) \), to map the feature matrix to a latent space for contrastive learning. Specifically, \( \bm{z}_{i}^{\alpha} = g_I(\bm{h}_i^{\alpha}) \) and \( \bm{z}_{i}^{\beta} = g_I(\bm{h}_i^{\beta}) \) are computed before evaluating the instance-level contrastive loss \cite{b15}. The pairwise similarity between embeddings is measured using cosine similarity:
\[
\scalebox{1}{
    \( s(\bm{z}_{i}^{\alpha}, \bm{z}_{j}^{\beta}) = \frac{
    \left( \bm{z}_i^{\alpha} \right)^T \left( \bm{z}_j^{\beta} \right) 
    }{
    \left\| \bm{z}_i^{\alpha} \right\|_2 \left\| \bm{z}_j^{\beta} \right\|_2
    } \),
}
\tag{22}
\label{22}
\]
where \( \alpha, \beta \in \{a, b\} \) and \( i, j \in [1, N] \).
For a given sample \( \bm{X}_{i}^a \), the instance-level contrastive loss is:
\[
\scalebox{1}{
    \(\ell_{i}^a = - \log \frac{\exp(s(\bm{z}_{i}^a, \bm{z}_{i}^b) / \tau_I)}{\sum_{j=1}^N \left[\exp(s(\bm{z}_{i}^a, \bm{z}_{j}^a) / \tau_I) + \exp(s(\bm{z}_{i}^a, \bm{z}_{j}^b) / \tau_I)\right]}\),
}
\tag{23}
\]
with \( \tau_I \) as the instance-level temperature parameter. The overall instance-level contrastive loss is averaged over all augmented samples:
\[
\scalebox{1}{
    \(\mathcal{L}_{INS} = \frac{1}{2N} \sum_{i=1}^N (\ell_{i}^a + \ell_{i}^b)\),
}
\tag{24}
\label{24}
\]
In JojoSCL, \(\mathcal{L}_{SURE}\) refines instance-level contrastive learning by aligning embeddings closer to their centroids, thus minimizing intra-cluster variance. This adjustment reduces the Euclidean distance between similar feature representations and increases it between dissimilar ones. As a result, \(\mathcal{L}_{SURE}\) improves cosine similarity for similar pairs and reduces it for dissimilar pairs:
\[
\scalebox{0.95}{$
\begin{aligned}
\left\| \bm{z}_i^{\alpha} - \bm{z}_j^{\beta} \right \|_2^2 &= \left\| \bm{z}_i^{\alpha} \right\|_2^2 + \left\| \bm{z}_j^{\beta} \right\|_2^2 - 2\cdot \left( \bm{z}_i^{\alpha}\right)^T \cdot \bm{z}_j^{\beta} \\
&= \left\| \bm{z}_i^{\alpha} \right\|_2^2 + \left\| \bm{z}_j^{\beta} \right\|_2^2 - 2\cdot \left\| \bm{z}_i^{\alpha} \right\|_2 \cdot \left\| \bm{z}_j^{\beta} \right\|_2 \cdot s(\bm{z}_{i}^{\alpha}, \bm{z}_{j}^{\beta}),
\end{aligned}
$}
\tag{25}
\]
which can be reformulated to show the inverse relationship:
\[
\scalebox{1}{$
s(\bm{z}_{i}^{\alpha}, \bm{z}_{j}^{\beta}) = \frac{
\left\| \bm{z}_i^{\alpha} \right\|_2^2 + \left\| \bm{z}_j^{\beta} \right\|_2^2 - \left\| \bm{z}_i^{\alpha} - \bm{z}_j^{\beta} \right \|_2^2
}{
2 \cdot \left\| \bm{z}_i^{\alpha} \right\|_2 \cdot \left\| \bm{z}_j^{\beta} \right\|_2
}.
$}
\tag{26}
\]
Thus, the integration of \(\mathcal{L}_{SURE}\) with instance-level contrastive learning not only enhances the alignment of embeddings with their centroids but also contributes to improved instance contrastive learning.

\subsection{Cluster-level Loss Formulation}

For cluster-level contrastive learning [14, 15], feature representations \(\{\bm{h}_i^a, \bm{h}_i^b \mid i=1,\ldots,N\}\) are projected into a \(K\)-dimensional space, where \(K\) represents the number of clusters. In this space, each component reflects the probability of belonging to a specific cluster. Let \(\bm{Y}^a \in \mathbb{R}^{N \times K}\) and \(\bm{Y}^b \in \mathbb{R}^{N \times K}\) denote the output matrices for the first and second augmentations, respectively, where \(\bm{Y}^a_{n,k}\) indicates the probability of the \(n\)-th sample belonging to cluster \(k\).

An MLP \( g_C(\cdot) \) transforms the feature matrix into a \(K\)-dimensional embedding space, resulting in cluster embeddings \(\bm{y}^a_i\) and \(\bm{y}^b_i\) for the \(i\)-th cluster under different augmentations. Positive pairs are formed by \(\{\bm{y}^{a_i}, \bm{y}^{b_i}\}\), while other pairs are treated as negative. The cosine similarity between cluster embeddings is computed as:
\[
\scalebox{1}{$
s(\bm{y}_i^{\alpha}, \bm{y}_j^{\beta}) = \frac{(\bm{y}_i^{\alpha})^T \bm{y}_j^{\beta}}{\| \bm{y}_i^{\alpha} \|_2 \| \bm{y}_j^{\beta} \|_2}
$},
\tag{27}
\label{27}
\]
where \( \alpha, \beta \in \{a, b\} \) and \( i, j \in [1, K] \). The loss for a cluster embedding \(\bm{y}^a_i\) is:
\[
\scalebox{1}{$
\hat{\ell}^a_i = - \log \frac{\exp(s(\bm{y}^a_i, \bm{y}^b_i)/\tau_C)}{\sum_{j=1}^K \left[ \exp(s(\bm{y}^a_i, \bm{y}^a_j)/\tau_C) + \exp(s(\bm{y}^a_i, \bm{y}^b_j)/\tau_C) \right]}
$},
\tag{28}
\label{28}
\]
where \(\tau_C\) is the temperature parameter. The total cluster-level contrastive loss is:
\[
\scalebox{1}{$
\mathcal{L}_{\text{CLU}} = \frac{1}{2K} \sum_{i=1}^K \left( \hat{\ell}^a_i + \hat{\ell}^b_i \right) - H(\bm{Y})
$},
\tag{29}
\label{29}
\]
where \( H(\bm{Y}) \) represents the entropy of the cluster probabilities:
\[
\scalebox{1}{$
H(\bm{Y}) = \sum_{i=1}^K \left[ P(\bm{y}^a_i) \log P(\bm{y}^a_i) + P(\bm{y}^b_i) \log P(\bm{y}^b_i) \right]
$},
\tag{30}
\label{30}
\]
with \( P(\bm{y}_{i}^\alpha) = {\sum_{j=1}^{N} \bm{Y}_{ji}^\alpha} / {\|\bm{Y}\|_1} \) for \( \alpha \in \{a,b \} \). This entropy term ensures well-distributed cluster assignments, preventing trivial solutions.

Moreover, \( \mathcal{L}_{\text{CLU}} \) benefits from \( \mathcal{L}_{\text{SURE}} \), which promotes tighter clustering around centroids and thus improves the accuracy and effectiveness of \( \mathcal{L}_{\text{CLU}} \).

\subsection{Final Loss Formulation}

The final loss function is the combination of the three loss functions (\ref{21}), (\ref{24}), and (\ref{29}) detailed in Section 3.3, 3.4, and 3.5:
\[
\scalebox{1}{$
\mathcal{L} = \mathcal{L}_{SURE} + \alpha \cdot \mathcal{L}_{INS} + \beta \cdot \mathcal{L}_{CLU},
$}
\tag{31}
\]
where \(\alpha\) and \(\beta\) are their parameters. 

\section{Experiments}
\subsection{Experimental Setup}

\begin{table}[b]
    \caption{Summary of datasets used in the study.}
    \centering
    \resizebox{0.48\textwidth}{!}{ 
    \begin{tabular}{llccc} 
        \toprule
        \textbf{Dataset}  & \textbf{Platform} & \textbf{\#Cells} & \textbf{\#Genes} & \textbf{\#Subtypes} \\ 
        \midrule
        \textbf{Adam}  & Drop-seq & 3660 & 23797 & 8 \\ 
        \textbf{Bladder}  & Microwell-seq & 2746 & 20670 & 16 \\ 
        \textbf{Chen}  & 10x & 12089 & 17550 & 46 \\ 
        \textbf{Human\_brain}  & Illumina MiSeq & 420 & 21609 & 8 \\ 
        \textbf{Klein}  & inDrop & 2717 & 24047 & 4 \\ 
        \textbf{Macosko}  & Drop-seq & 14653 & 11422 & 39 \\ 
        \textbf{Mouse}  & Microwell-seq & 2100 & 20670 & 16 \\ 
        \textbf{Shekhar}  & Drop-seq & 27499 & 13166 & 19 \\ 
        \textbf{Yan}  & Tang & 90 & 16383 & 7 \\ 
        \textbf{10X\_PBMC}  & 10x & 4271 & 16653 & 8 \\ 
        \bottomrule
    \end{tabular}
    }

    \label{tab:datasets}
\end{table}

\begin{table*}[t]
    \caption{Clustering performance of different models across various datasets, based on 10 consecutive runs, is evaluated in terms of ARI and NMI. The best clustering result for each dataset is bolded, and the second-best result is underlined.}
    \centering
    \renewcommand{\arraystretch}{1.2} 
    \setlength{\tabcolsep}{4pt} 
    \begin{tabular}{l@{\hskip 5pt}>{\centering\arraybackslash}p{0.82cm}>{\centering\arraybackslash}p{0.82cm}|>{\centering\arraybackslash}p{0.82cm}>{\centering\arraybackslash}p{0.82cm}|>{\centering\arraybackslash}p{0.82cm}>{\centering\arraybackslash}p{0.82cm}|>{\centering\arraybackslash}p{0.82cm}>{\centering\arraybackslash}p{0.82cm}|>{\centering\arraybackslash}p{0.82cm}>{\centering\arraybackslash}p{0.82cm}|>{\centering\arraybackslash}p{0.82cm}>{\centering\arraybackslash}p{0.82cm}}
        \toprule
        \textbf{Dataset} & \multicolumn{2}{c|}{\textbf{Seurat}} & \multicolumn{2}{c|}{\textbf{scziDesk}} & \multicolumn{2}{c|}{\textbf{scDeepCluster}} & \multicolumn{2}{c|}{\textbf{Contrastive-sc}}  & \multicolumn{2}{c|}{\textbf{ScCCL}} & \multicolumn{2}{c}{\textbf{JojoSCL}} \\
        \midrule
        \textbf{Metrics} & \textbf{ARI} & \textbf{NMI} & \textbf{ARI} & \textbf{NMI} & \textbf{ARI} & \textbf{NMI} & \textbf{ARI} & \textbf{NMI} & \textbf{ARI} & \textbf{NMI} & \textbf{ARI} & \textbf{NMI} \\
        \midrule
        \textbf{Adam}        & 0.6806 & 0.7151 & 0.8273 & 0.8340& 0.7892 & 0.7691 & 0.9034 & 0.8973 & \underline{0.9133}& \underline{0.9008}& \textbf{0.9343}& \textbf{0.9191}\\
        \textbf{Bladder}     & 0.5825 & 0.6310 & 0.4907 & 0.6051 & \underline{0.6030}& \underline{0.7370}& 0.5546 & 0.6704 & 0.5798 & 0.7332 & \textbf{0.6079}& \textbf{0.7507}\\
        \textbf{Chen}        & 0.5907 & 0.5563 & \underline{0.7651}& 0.6413 & 0.3791 & 0.3069 & 0.7224 & \underline{0.6810}& {0.7646}& 0.6802 & \textbf{0.8168}& \textbf{0.7362}\\
        \textbf{Human\_brain} & 0.7671 & 0.7315 & 0.8330 & 0.8328 & 0.8215 & 0.8007 & 0.8306 & 0.8179 & \underline{0.8565}& \underline{0.8340}& \textbf{0.8905}& \textbf{0.8510}\\
        \textbf{Klein}       & 0.7436 & 0.7275 & \underline{0.8014}& \underline{0.7883}& 0.7837 & 0.7512 & 0.6772 & 0.6559 & 0.7835 & 0.7745 & \textbf{0.8892}& \textbf{0.8547}\\
        \textbf{Macosko}     & 0.6335 & 0.7720 & 0.7252 & \textbf{0.8247}& 0.6209 & 0.7931 & 0.7762 & 0.7917 & \underline{0.8581}& 0.7985& \textbf{0.8614}& \underline{0.8145}\\
        \textbf{Mouse}       & 0.6277 & 0.6641& \underline{0.7859}& \underline{0.8013}& \textbf{0.8177}& \textbf{0.8318}& 0.7210 & 0.7554 & 0.6400& 0.7033 & 0.6631& 0.6995
\\
        \textbf{Shekhar}     & 0.7106 & 0.8377 & 0.5651 & 0.6426 & 0.6796 & 0.7995 & 0.7050& 0.8341 & \underline{0.9552}& \underline{0.8860}& \textbf{0.9624}& \textbf{0.8997}\\
        \textbf{Yan}         & 0.7095 & 0.7644 & \underline{0.8665}& 0.8713 & 0.8109 & 0.8663 & 0.8596 & 0.8710 & \textbf{0.8744}& \textbf{0.8813}& 0.8662& \underline{0.8793}\\
        \textbf{10X\_PBMC}   & 0.5316 & 0.7129 & 0.6488 & 0.7366 & 0.7640& 0.7580& 0.7644 & 0.7569 & \underline{0.7866}& \underline{0.7782}& \textbf{0.8080}& \textbf{0.8025}\\
        \bottomrule
 \textbf{Average}& 0.6577& 0.7112& 0.7309& 0.7578& 0.7070& 0.7414& 0.7514& 0.7732& 0.8012& 0.7970& 0.8300&0.8207\\
 \bottomrule
    \end{tabular}
    \label{tab:performance_comparison}
\end{table*}

\noindent \textbf{Dataset:} We evaluate our method on ten datasets from various platforms to compare its performance against different models. Summary statistics for these datasets are provided in Table 1.

\noindent \textbf{Competing models:} Our model was evaluated against five leading scRNA-seq clustering models: Seurat \cite{b7}, scziDesk \cite{b35}, scDeepCluster \cite{b13}, Contrastive-sc \cite{b17}, and ScCCL \cite{b22}. These models represent a diverse array of approaches to scRNA-seq clustering. Specifically, Seurat is built on graph-based clustering, while scziDesk and scDeepCluster are deep clustering models. We also compare JojoSCL with other contrastive learning models, including Contrastive-sc and ScCCL.

\noindent \textbf{Evaluation metrics:}
We use two widely-adopted metrics to evaluate clustering performance: the Adjusted Rand Index (ARI) and the Normalized Mutual Information (NMI). Higher values of ARI and NMI indicate superior clustering performance.

\noindent \textbf{Computational complexity:} The computational complexity for JojoSCL is \(
O(E \cdot (N^2))
\), with \(E\) the number of training epochs and \(N\) the batch size.

\subsection{Comparison results}

Table \ref{tab:performance_comparison} presents the clustering performance results for JojoSCL and the five competing methods on the datasets from Table \ref{tab:datasets}. 
Based on the results, we observe:

\noindent 1. JojoSCL, consistently outperforms all other methods in ARI and NMI across 9 out of 10 datasets. Specifically, it achieves an average of 26\% higher ARI and 15\% higher NMI compared to Seurat. JojoSCL also surpasses deep clustering methods, scziDesk and scDeepCluster, by 15\% in ARI and 9\% in NMI, and previous contrastive clustering methods, Contrastive-sc and ScCCL, by 7\% in ARI and 5\% in NMI.

\begin{figure}[t]
    \centering
    \subfigure[Shekhar]{
        \includegraphics[width=0.148\textwidth]{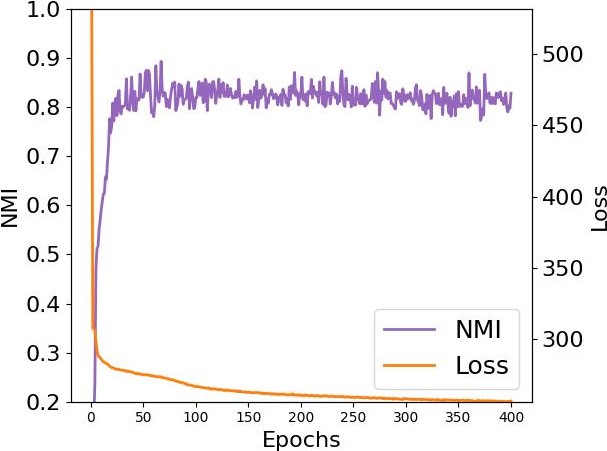}
        \label{fig:shekhar_NMI}
    }\hfill
    \subfigure[Human\_brain]{
        \includegraphics[width=0.148\textwidth]{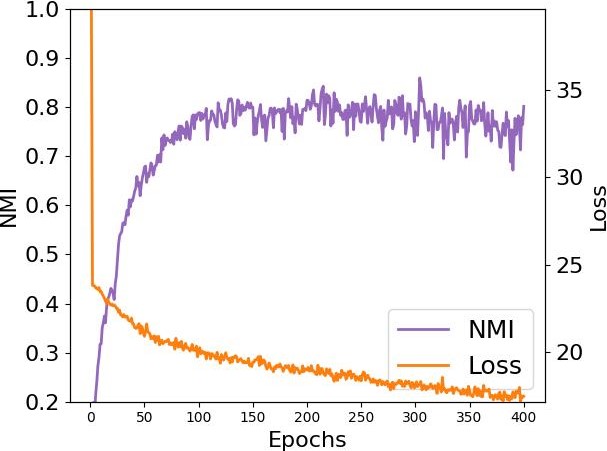}
        \label{fig:human_brain_NMI}
    }\hfill
    \subfigure[Bladder]{
        \includegraphics[width=0.148\textwidth]{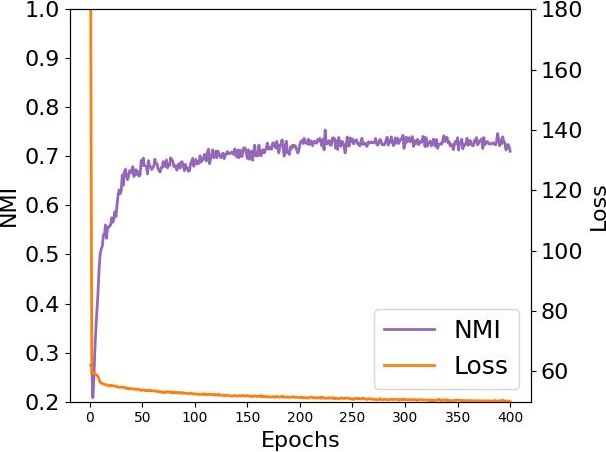}
        \label{fig:Bladder_NMI}
    }
    \caption{The convergence of NMI and loss across 400 epochs for JojoSCL on Shekhar, human\_brain, and Bladder datasets.  For some datasets, the NMI peaked and then showed signs of overfitting, while for others, stable convergence persisted beyond 400 epochs. }
    \label{fig:NMI_Loss}
\end{figure}

\noindent 2. In datasets like Adam, Human\_brain, and 10X\_PBMC, where other models perform well, JojoSCL shows incremental improvements and achieves the best results. For instance, in the Adam dataset, JojoSCL has an ARI of 0.9343 and an NMI of 0.9191, surpassing the second-best results by 0.0210 in ARI and 0.0183 in NMI. JojoSCL also excels in datasets where contrastive methods typically lag behind deep clustering, such as Klein, outperforming the second-best model (scziDesk) by 0.0878 in ARI and 0.0664 in NMI.

\noindent 3. On the Chen dataset with 46 subtypes, JojoSCL achieves an ARI of 0.8168 and an NMI of 0.7362, significantly surpassing other models. In the Shekhar dataset with 19 subtypes, JojoSCL outperforms leading deep clustering methods by 54\% in ARI and 25\% in NMI, and demonstrates progress over contrastive clustering models that have shown promising performance.

\noindent 4. JojoSCL demonstrates significantly faster convergence and reduced overall training time. We conduct an analysis to evaluate the convergence behavior of the proposed model on the Shekhar, Human\_brain, and Bladder datasets. As shown in Fig. \ref{fig:NMI_Loss}, our method achieves stable clustering and convergence within a few training epochs, indicating its efficiency and effectiveness.

\begin{table}[t]
    \caption{Comparison of JojoSCL's clustering performance on datasets with and without noise in NMI.}
    \centering
    \resizebox{\linewidth}{!}{%
        \begin{tabular}{lccc}
            \toprule
            \textbf{Dataset} & \textbf{With Noise} & \textbf{Without Noise} & \textbf{Difference} \\
            \midrule
            \textbf{Adam}        & \textbf{0.9191}& 0.9029 & 0.0162 \\
            \textbf{Bladder}     & \textbf{0.7507}& 0.7399 & 0.0108 \\
            \textbf{Chen}        & \textbf{0.7362}& 0.7044 & 0.0318 \\
            \textbf{Human\_brain} & 0.8510  & \textbf{0.8563}& -0.0053 \\
            \textbf{Klein}       & 0.8447 & \textbf{0.8551}& -0.0104 \\
            \textbf{Macosko}     & \textbf{0.8145}& 0.7932 & 0.0213 \\
            \textbf{Mouse}       & 0.6995 & \textbf{0.7213}& -0.0218 \\
            \textbf{Shekhar}     & \textbf{0.8997}& 0.8893 & 0.0104 \\
            \textbf{Yan}         & \textbf{0.9070}& 0.8876 & 0.0194 \\
            \textbf{10X\_PBMC}   & \textbf{0.8025}& 0.7920  & 0.0105 \\
            \bottomrule
        \end{tabular}%
    }
    \label{tab:noise_comparison}
\end{table}

\subsection{Robustness analysis with noise conditions}

As discussed in Section 3.2, we address scRNA-seq data challenges using a masking strategy where gene expression values are set to zero and Gaussian noise is added during data augmentation. To assess the impact of noise on JojoSCL’s performance, we compared results with and without noise, keeping other conditions constant. The findings are summarized in Table \ref{tab:noise_comparison}.

\begin{figure}[b]
    \centering
    \subfigure[NMI: \(0.7967\)]{
        \includegraphics[width=0.148\textwidth]{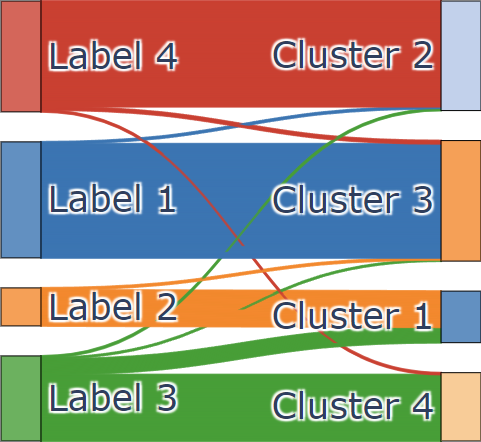}
        \label{fig:subfig1}
    }\hfill
    \subfigure[NMI: \(0.8217\)]{
        \includegraphics[width=0.148\textwidth]{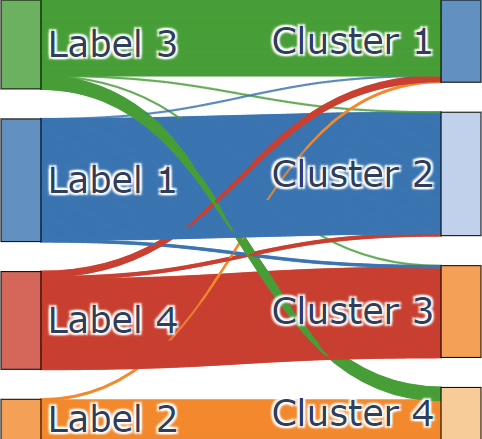}
        \label{fig:subfig2}
    }\hfill
    \subfigure[NMI: \(0.7935\)]{
        \includegraphics[width=0.148\textwidth]{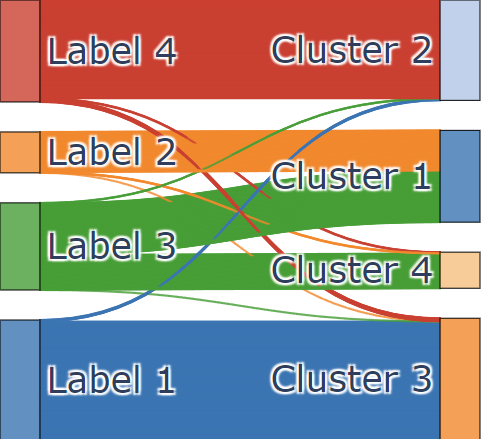}
        \label{fig:subfig3}
    }
    \caption{Clustering results for the Klein dataset using JojoSCL with random data drops of (a) 20\%, (b) 50\%, and (c) 80\% measured by NMI.}
    \label{fig:rand_downsample_Klein}
\end{figure}

The results show that noise contributes to better clustering performances in 7 of 10 datasets, notably in datasets with many cell types (e.g., Chen with 46 subtypes and Macosko with 39 subtypes) and in smaller datasets (e.g., Yan with 90 cells). 

A two-sided paired t-test comparing the NMI with and without noise across all datasets yielded a t-statistic of 1.618 and a p-value of 0.140. This indicates that we cannot reject the null hypothesis at the 5\% or 10\% significance levels, suggesting that JojoSCL performs robustly and effectively even without noise. 

\subsection{Robustness analysis with partial data}

Downsampling is a common method to test model performance on smaller or incomplete datasets. In scRNA-seq clustering, it evaluates robustness with limited or imbalanced data. We applied downsampling to the Klein dataset, which has uneven cell type distributions, using both random and stratified methods. Random downsampling removes data indiscriminately, while stratified downsampling maintains proportional cell type representation.

\begin{figure}[t]
    \centering
    \subfigure[NMI: \(0.8167\)]{
        \includegraphics[width=0.148\textwidth]{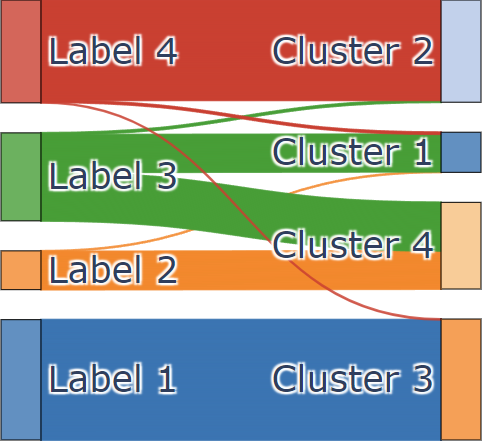}
        \label{fig:subfig4}
    }\hfill
    \subfigure[NMI: \(0.8538\)]{
        \includegraphics[width=0.148\textwidth]{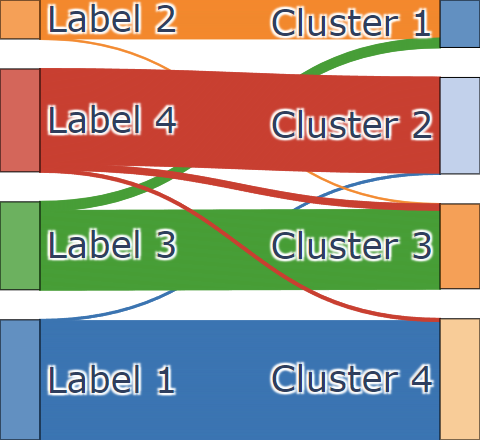}
        \label{fig:subfig5}
    }\hfill
    \subfigure[NMI: \(0.8230\)]{
        \includegraphics[width=0.148\textwidth]{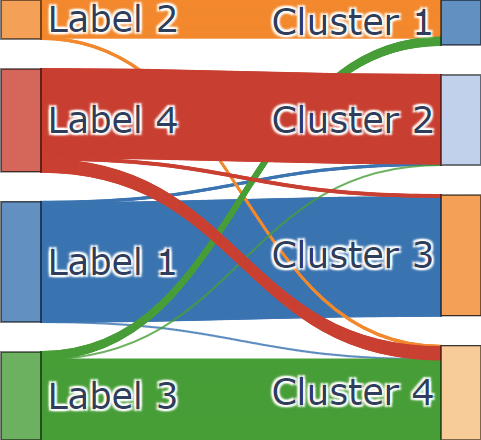}
        \label{fig:subfig6}
    }
    \caption{Clustering results for the Klein dataset using JojoSCL with stratified data drops of (a) 20\%, (b) 50\%, and (c) 80\% measured by NMI.}
    \label{fig:fixed_downsample_Klein}
\end{figure}

Clustering results for various downsampling rates are shown in Fig. \ref{fig:rand_downsample_Klein} and Fig. \ref{fig:fixed_downsample_Klein}. Performance declines with reduced dataset size: NMI drops from 0.8167 to 0.7967 at 20\%, from 0.8538 to 0.8217 at 50\%, and from 0.8230 to 0.7935 at 80\%. Despite these decreases, JojoSCL's NMI remains higher than competing methods across all levels. Additionally, JojoSCL effectively identifies most samples for each cell type, demonstrating its stability and robustness.

\begin{figure}[b]
    \centering
    \includegraphics[width=\linewidth]{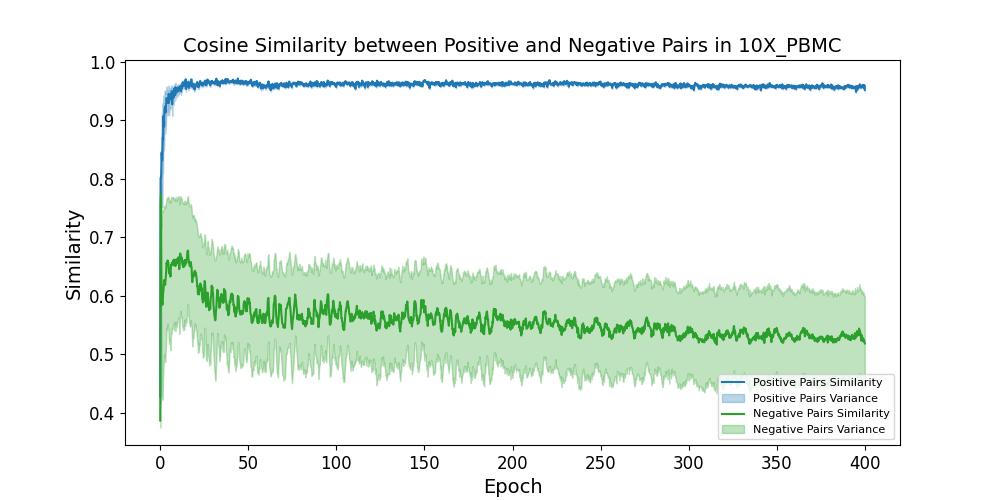}
    \caption{ The separation between positive and negative pairs and the growing difference observed as JojoSCL undergoes additional training epochs with the 10X\_PBMC dataset.}
    \label{fig: cosine similarity}
\end{figure}

\begin{table}[t]
\caption{The mean and variance of the difference in cosine similarity \( s(z_{i}^{\alpha}, z_{j}^{\beta}) \) between positive and negative pairs with and without \(\mathcal{L}_\text{SURE}\) over 400 training epochs.}
\centering
\scriptsize
\resizebox{\linewidth}{!}{
\begin{tabular}{l|cc|cc}
\toprule
\textbf{Dataset} & \multicolumn{2}{c|}{\textbf{With } \(\mathbf{\mathcal{L}_{SURE}}\)} & \multicolumn{2}{c}{\textbf{Without } \(\mathbf{\mathcal{L}_{SURE}}\)} \\
\midrule
\textbf{Metrics} & \textbf{Mean} & \textbf{Variance} & \textbf{Mean} & \textbf{Variance} \\
\midrule
\textbf{Adam} & \textbf{0.3944}& 0.0010 
& 0.3898 & 0.0006
\\
\textbf{Bladder} & \textbf{0.3267}& 0.0029 
& 0.3182 & 0.0013
\\
\textbf{Chen} & \textbf{0.4069}& 0.0012 
& 0.3812 & 0.0007
\\
\textbf{Human\_brain} & \textbf{0.4152}& 0.0038 
& 0.3993 & 0.0039
\\
\textbf{Klein} & \textbf{0.4336}& 0.0011 
& 0.4275 & 0.0011
\\
\textbf{Macosko} & \textbf{0.4443}& 0.0008 
& 0.4315 & 0.0012
\\
\textbf{Mouse} & 0.3778 & 0.0007 
& \textbf{0.3874}& 0.0007
\\
\textbf{Shekhar} & \textbf{0.4302}& 0.0009 
& 0.4190 & 0.0005
\\
\textbf{Yan} & \textbf{0.4993}& 0.0022 
& 0.4839 & 0.0024
\\
\textbf{10X\_PBMC} & \textbf{0.3868}& 0.0004 
& 0.3740 & 0.0004
\\
\midrule
\textbf{Average} & \textbf{0.4115}& 0.0015 & 0.4012 & 0.0013
\\
\bottomrule
\end{tabular}
}
\label{tab:similarity_metrics}
\end{table}

\begin{figure}[b]
    \centering
    \includegraphics[width=\linewidth]{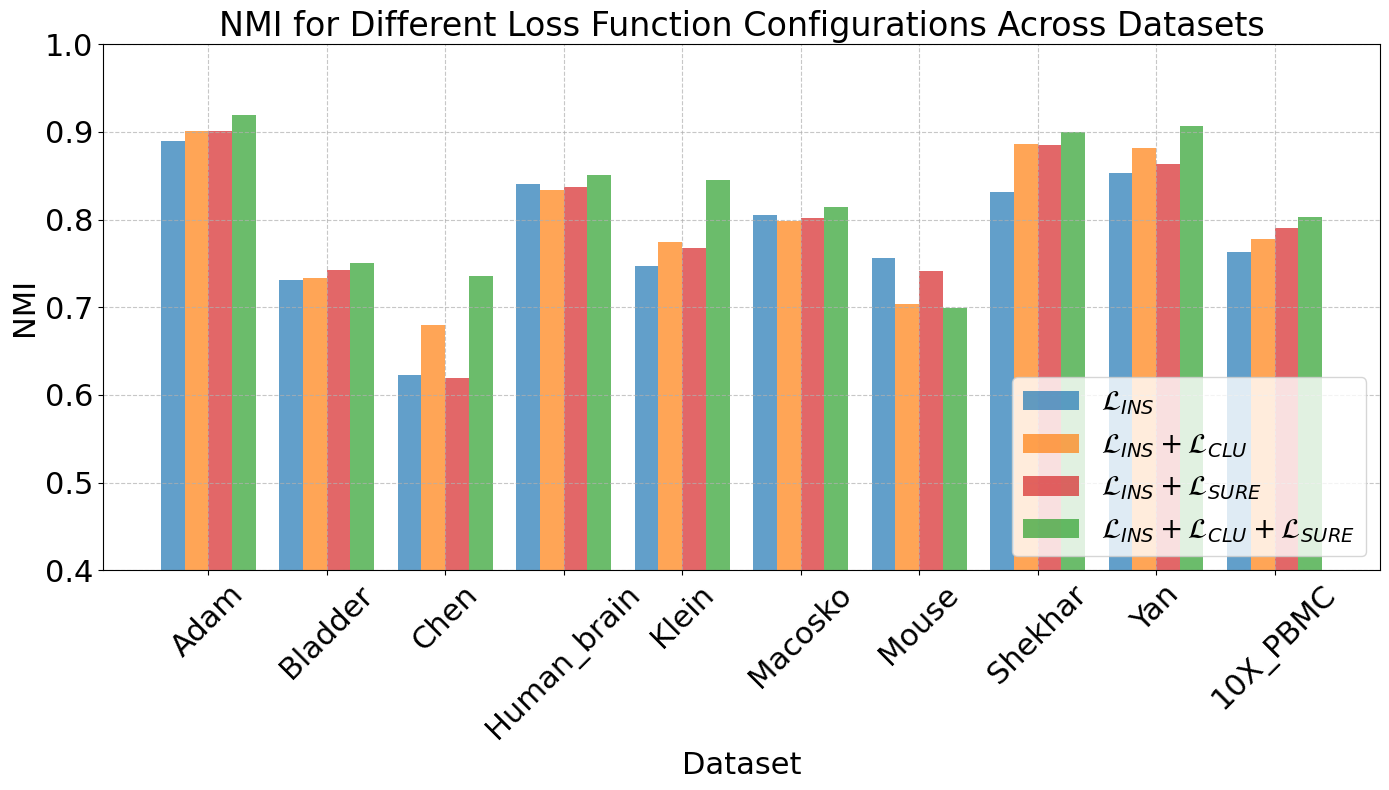}
    \caption{Different combinations of \(\mathcal{L}_{SURE}\), \(\mathcal{L}_{INS}\), and \textbf{\(\mathcal{L}_{CLU}\)} with their corresponding clustering performance, measured in NMI, on datasets listed in Table \ref{tab:datasets}.}
    \label{fig: performance enhancement}
\end{figure}

\subsection{Ablation Studies: Impact of \texorpdfstring{\(\mathcal{L}_{SURE}\)}{L\_SURE} on the pairwise similarity}

We assess the effect of \(\mathcal{L}_{SURE}\) on pairwise similarity by comparing JojoSCL with and without \(\mathcal{L}_{SURE}\). When \(\mathcal{L}_{SURE}\) is omitted, only \(\mathcal{L}_{INS}\) and \(\mathcal{L}_{CLU}\) are used. Results in Table \ref{tab:similarity_metrics} show that \(\mathcal{L}_{SURE}\) significantly improves the difference in pairwise similarity, enhancing contrastive learning effectiveness. Specifically, \(\mathcal{L}_{SURE}\) increases the mean of the difference in cosine similarity between positive and negative pairs in 9 out of 10 datasets. A two-sided paired t-test was conducted to compare mean differences in cosine similarity with and without $\mathcal{L}_{SURE}$. The test yielded a t-statistic of 3.5648 and a p-value of 0.0061, indicating that the greater mean difference in cosine similarity between positive and negative pairs with $\mathcal{L}_{SURE}$ compared to without $\mathcal{L}_{SURE}$ is statistically significant.

\subsection{Ablation Studies: Performance enhancement by \texorpdfstring{\(\mathcal{L}_{SURE}\)}{L-SURE}}

To validate the clustering performance enhancement attributed to $\mathcal{L}_{SURE}$, we evaluate four variants of our method with different combinations of loss functions across all datasets. The results are presented in Fig. \ref{fig: performance enhancement}. 

The combination of all three loss functions in the full model JojoSCL achieved the highest performance in 9 out of 10 datasets, while the model using only $\mathcal{L}_{INS}$ exhibited the lowest clustering performance. Our theoretical analysis in Sections 3.4 and 3.5 suggests that $\mathcal{L}_{SURE}$ effectively refines the process of distinguishing between instances by bringing similar ones closer to the centroids and improving cluster separation. These findings align with our theoretical expectations, as the combination of $\mathcal{L}_{INS} + \mathcal{L}_{SURE}$ demonstrated the second-best performance, which indicates that $\mathcal{L}_{SURE}$ contributes to performance improvement beyond $\mathcal{L}_{CLU}$ under certain conditions.

\section{Conclusion}

In this paper, we have developed a novel self-supervised contrastive learning framework for scRNA-seq clustering tasks. Our approach introduces a new shrinkage estimator based on hierarchical Bayesian estimation and regulated by Stein's Unbiased Risk Estimate. We demonstrate that this shrinkage method practically enhances both instance-level and cluster-level contrastive learning, improving the model’s ability to address the challenges of high dimensionality and sparsity in scRNA-seq clustering.
Experiments on ten scRNA-seq datasets show that our method significantly outperforms competing methods. Further validation through robustness analysis and ablation studies confirms the effectiveness of our approach.

\end{document}